# An a Priori Exponential Tail Bound for $k$-Folds Cross-Validation

**Karim Abou-Moustafa**
Dept. of Computing Science
University of Alberta
Edmonton, AB T6G 2E8, Canada
aboumous@ualberta.ca

**Csaba Szepesvári**
Department of Computing Science
University of Alberta
Edmonton, AB T6G 2E8, Canada
szepesva@ualberta.ca

## Abstract

We consider a priori generalization bounds developed in terms of cross-validation estimates and the stability of learners. In particular, we first derive an exponential Efron-Stein type tail inequality for the concentration of a general function of $n$ independent random variables. Next, under some reasonable notion of stability, we use this exponential tail bound to analyze the concentration of the $k$-fold cross-validation (KFCV) estimate around the true risk of a hypothesis generated by a general learning rule. While the accumulated literature has often attributed this concentration to the bias and variance of the estimator, our bound attributes this concentration to the stability of the learning rule and the number of folds $k$. This insight raises valid concerns related to the practical use of KFCV, and suggests research directions to obtain reliable empirical estimates of the actual risk.

## 1 Introduction

$k$-Folds cross-validation (KFCV) is perhaps the most prominent procedure for risk estimation [Stone, 1974, Geisser, 1975]. It is used in practice with the promise of being more accurate than the training error (also known as the resubstitution estimate), while avoiding the high computational cost associated with the deleted estimate (which is also known as the leave-one-out estimate). An important problem is to understand the concentration of the KFCV around the true (random) risk and it is this problem that we investigate in this paper.

Various works have considered different aspects of this question. Blum et al. [1999] suggest to replace the hypothesis that uses all samples with a hypothesis randomly selected from the $k$ hypotheses that are obtained while running KFCV. Noting that the average of the risks if these $k$ (random) hypotheses has the same risk on expectation than the risk of a randomly picked hypothesis of these $k$ hypotheses, they observe that the deviation of the KFCV from the average of the risks concentrates at least as well as the holdout estimate obtained on one of the folds concentrates around the risk of a hypothesis produced while training on the remaining $k-1$ folds. While this observation combined with stability arguments can be used to derive a concentration bound of the deviation of interest, the resulting concentration result is suboptimal in the regime when $k$ is comparable with $n$, the number of training examples (this is the regime of the deleted estimate). In particular, one term of the expected squared deviation behaves (closely related to the variance) as $k/n$, which leads to a vacuous bound for $k = n$. Kale et al. [2011], using a version of the so-called Efron-Stein inequality (due to Steele), show a bound on the variance $V_{\mathrm{KFCV}}$ of KFCV in terms of the variance $V$ of the holdout over one fold. In particular, they show that $V_{\mathrm{KFCV}} \leq V/k + (1 - 1/k)\beta\sqrt{V/2}$ where $\beta$ is the (replacement-type) $L^2$-stability coefficient of the learning rule. When $V \leq k/n$ and $\beta = 1/(k^{1/2}n)$ (as is expected for some typical learning rules and learning settings under proper normalization), $V_{\mathrm{KFCV}} \leq 2/n$, showing that under favorable conditions the variance of KFCV is not adversarially impacted by the choice of $k$ in the full spectrum of $k = 2$ to $k = n$. Cornec [2017], in the spirit of sanity-check bounds, shows that for empirical risk minimizers over VC-classes, the worst-case error of the KFCV estimate is not much worse than that of the training error, a result that was proven for stable learning rules by Kearns and Ron [1999].

In this work we consider the exponential concentration of the KFCV estimate around the actual risk of a hypothesis returned by a stable learning rule under distribution-dependent notions of stability. Earlier works have derived such concentration results for the *deleted estimate*, and learning rules that are *uniformly stable* in the sense that no matter how the input to the learning rule is selected, and no matter what value is used as a test example, replacing (or removing) one example in the input, the prediction loss will change only in a limited fashion [Bousquet and Elisseeff, 2002]. The stability coefficient of a learning rule is the amount of this change. Bousquet and Elisseeff [2002] considered the concentration of the right tail of the difference between the deleted estimate and the (random) risk of a hypothesis returned by a uniformly stable learning rule and find that this is comparable to the concentration of the right tail of the deviation between the resubstitution estimate and the true random risk, adjusted for the stability of the learning rule. The main observation of Bousquet and Elisseeff [2002] is that uniform stability (a worst-case notion over all training and test examples) allows an elegant use of McDiarmid's inequality, which leads to exponential tail bounds. Kutin and Niyogi [2002] and Rakhlin et al. [2005] consider a softening of the stringent requirement underlying uniform stability to "almost everywhere" stability. Under the new notion, the uniform stability requirement needs to hold with some probability that can be less than one by a positive amount $\delta$. On the event when the uniform requirement does not hold, the magnitude of the loss must be uniformly bounded. The tail bound they derive is the sum of an exponential term and a term that depends on the failure probability $\delta$. While Kutin and Niyogi [2002] prove their result by extending McDiarmid's inequality, Rakhlin et al. [2005] used the higher-moment version of the Efron-Stein inequality due to Boucheron et al. [2003] (the usefulness of the classic Efron-Stein inequality in connection to generalization bounds for stable algorithms was already noticed by Bousquet and Elisseeff [2002], and later, as noted above, was used by Kale et al. [2011]).

Uniform stability is unpleasantly restrictive: Unlike other notions of stability (e.g., $L^2$, or $L^1$ stability), it is insensitive to the data-generating distribution. This is problematic as it removes the possibility of studying large classes of learning rules, or even classes of problems. One particularly striking example is binary classification with the zero-one loss. For this problem, as it was noted already by Bousquet and Elisseeff [2002], *no nontrivial algorithm* can be uniformly $\beta$-stable with $\beta < 1$ [Bousquet and Elisseeff, 2002]. Another example when uniform stability fails is regression with unbounded response variables and losses. In addition, as noted earlier, uniform stability is distribution-free and is thus unsuitable to studying finer details of learning. As due to the no-free lunch results, for sufficiently "rich" learning settings, no single learning rule can be expected to perform uniformly well, there is much to be gained by studying the biases of learning rules.

What notions of stability is then suitable for deriving appropriate distribution dependent concentration bounds for KFCV? Since we are interested in the tail properties of KFCV and higher moments are sufficient and necessary to characterize the tails of random variables, it is natural to expect that the whole family of $L^q$-stability coefficients with $q \geq 1$ would play a role in determining the tail behavior of KFCV. The advantage of using $L^q$ stability coefficients to uniform (which in a way are close to $L^\infty$ coefficients) is that they are distribution dependent and are nontrivial even when the uniform stability coefficient is uncontrolled. Recent, yet unpublished work by Celisse and Guedj [2016] indeed demonstrated that the family of $L^q$ stability coefficients can be successfully used to study the deviation of *deleted estimates*. While we also use the same family of stability coefficients, our work goes beyond the work of Celisse and Guedj [2016] in that we consider *for the first time in the literature* distribution dependent concentration bounds for the KFCV estimate. While our techniques resemble those of Celisse and Guedj [2016], we streamline several steps of their proofs. One difference is that we build directly on the elegant Efron-Stein style exponential inequality of Boucheron et al. [2003], while Celisse and Guedj [2016] chose a different route.

## 2 Setup and Notations

We consider learning in Vapnik's framework for risk minimization with bounded losses [Vapnik, 1995]: A learning problem is specified by the triplet $(\mathcal{H}, \mathcal{X}, \ell)$, where $\mathcal{H}, \mathcal{X}$ are sets and $\ell : \mathcal{H} \times \mathcal{X} \to [0, 1]$. The set $\mathcal{H}$ is called the *hypothesis space*, $\mathcal{X}$ is called the *instance space*, and $\ell$ is called the *loss function*. The loss $\ell(h, x)$ indicates how well a hypothesis $h$ explains (or fits) an instance $x \in \mathcal{X}$.

The learning problem is defined as follows: A learner $\mathtt{A}$ sees a sample in the form of a sequence $\mathcal{S}_n = (X_1, \ldots, X_n) \in \mathcal{X}^n$ where $(X_i)_i$ is sampled in an independent and identically distributed (*i.i.d*) fashion from some unknown distribution $\mathscr{P}$ and returns a hypothesis $\widehat{h}_n = \mathtt{A}(\mathcal{S}_n) \in \mathcal{H}$ based solely on $X_1, \ldots, X_n$.[1] The goal of the learner is to pick hypotheses with a small *risk* (defined shortly). For readers familiar with learning theory we remark that as opposed to most of statistical learning theory, the only role $\mathcal{H}$ plays is to collect the universe of all choices available to learning rules. In particular, unlike in most of the literature on statistical learning theory, it will not be used to "control the bias of learners".

We assume that a learner is able to process data of different cardinality. Hence, a learner will be identified with a map $\mathtt{A} : \cup_n \mathcal{X}^n \to \mathcal{H}$. We only consider deterministic learning rules in this work; the extension to randomizing learning rules is left for future work. Given a distribution $\mathscr{P}$ on $\mathcal{X}$, the risk of a *fixed hypothesis* $h \in \mathcal{H}$ is defined to be $R(h, \mathscr{P}) = \mathbb{E}\left[\ell\left(h, X\right)\right]$, where $X \sim \mathscr{P}$. Since $\mathcal{S}_n$ is random, so are $\mathtt{A}(\mathcal{S}_n)$ and $R(\mathtt{A}(\mathcal{S}_n), \mathscr{P})$, the latter of which can be also written as $\mathbb{E}[\ell\left(\mathtt{A}(\mathcal{S}_n), X\right) | \mathcal{S}_n]$, where $X \sim \mathscr{P}$ is independent of $\mathcal{S}_n$. Ideal learners keep the risk $R(\mathtt{A}(\mathcal{S}_n), \mathscr{P})$ of the hypothesis returned by $\mathtt{A}$ "small" for a wide range of distributions $\mathscr{P}$.

## 2.1 Quality Assessment of Learners

Most of statistical learning theory is devoted to answering the following two questions:

1. *A posteriori* performance assessment: How well *did* $\mathtt{A}$ work on some data $\mathcal{S}_n$ drawn from some distribution $\mathscr{P}$?
2. *A priori* performance prediction: How well *will* $\mathtt{A}$ perform on data $\mathcal{S}_n$ that will be drawn from some distribution $\mathscr{P}$?

For both questions, the answer should be given in terms of the risk $R(\mathtt{A}(\mathcal{S}_n), \mathscr{P})$ of the hypothesis $\mathtt{A}(\mathcal{S}_n)$. Since $\mathcal{S}_n$ and $\mathtt{A}(\mathcal{S}_n)$ are random quantities, in general, the answers to the above questions will be upper bounds, the so-called *generalization bounds*, on the random risk $R(\mathtt{A}(\mathcal{S}_n), \mathscr{P})$ that have a probabilistic nature; i.e. the bounds hold with high probability, or hold for the expected risk $R_n(\mathtt{A}, \mathscr{P}) = \mathbb{E}[\ell\left(\mathtt{A}(\mathcal{S}_n), X\right)]$, or the higher moments of the risk.

The two questions are similar in that both of them are concerned with performance on unseen data (since the definition of the risk involves future unseen data). As a result, often the questions are answered using similar tools. The two questions are also fundamentally different: in the case of the first question the data $\mathcal{S}_n$ that produces the hypothesis $\mathtt{A}(\mathcal{S}_n)$ is already given, while in the second case the data is yet unknown at the time when the question is asked. Correspondingly, we call bounds answering the first question *a posteriori* ("after the fact") bounds, while we call bounds answering the second question *a priori* bounds. Ideal a posteriori bounds depend on $\mathtt{A}$, $\mathscr{P}$, and $\mathcal{S}_n$ (i.e., these bounds should be learner-dependent, distribution-dependent, and data-dependent), while in the case of a priori bounds, the bound can at best depend on $\mathtt{A}$ and $\mathscr{P}$ (i.e., they can be learner- and distribution-dependent).[2]

In this paper we consider the second question, i.e., a priori generalization bounds. In particular, we consider a priori generalization bounds, but not for assessing the performance of particular learners, but to assess the performance of a general performance assessment tool, cross-validation estimation in relation to its ability to deliver good risk estimates for a host of learners. A very interesting avenue for further research, which we hope could be built on our results, is to develop a posteriori generalization bounds for cross-validation estimation.

## 2.2 Risk Estimators

The generalization bounds on the risk usually center on some point-estimate of the random risk $R(\mathtt{A}(\mathcal{S}_n), \mathscr{P})$. Many estimators are based on calculating the sample mean of losses in one form or another. For any fixed hypothesis $h \in \mathcal{H}$ and data $\mathcal{S}_n$, the sample mean of losses of $h$ against $\mathcal{S}_n$, also known as the *empirical risk* of $h$ on $\mathcal{S}_n$, is given by $\widehat{R}(h, \mathcal{S}_n) = \frac{1}{n}\sum_{i=1}^{n} \ell(h, X_i)$. Plugging $\mathtt{A}(\mathcal{S}_n)$ into $\widehat{R}(\cdot, \mathcal{S}_n)$ we get the *training error* or *resubstitution (RES) estimate* [Devroye and Wagner, 1979]: $\widehat{R}_{\text{RES}}(\mathtt{A}, \mathcal{S}_n) = \widehat{R}(\mathtt{A}(\mathcal{S}_n), \mathcal{S}_n)$. The resubstitution estimate is often overly "optimistic", i.e., it underestimates the actual risk $R(\mathtt{A}(\mathcal{S}_n), \mathscr{P})$.

[1] The set $\mathcal{X}$ is thus measurable. In general, for the sake of minimizing clutter, we will skip mentioning measurability issues; in particular, all functions are assumed to be measurable as needed.

[2] A large part of statistical learning theory is devoted to developing distribution-free bounds (i.e., distribution independent, subject to some "minor" restrictions on $\mathscr{P}$) as these bounds can be used to assess the worst-case performance of a learner over some large class of distributions. Distribution dependent bounds often but not always lead to such distribution-free bounds.

The *leave–one–out* or *deleted (DEL) estimate* [Devroye and Wagner, 1979], defined as $\widehat{R}_{\text{DEL}}(\mathtt{A}, \mathcal{S}_n) = \frac{1}{n}\sum_{i=1}^{n} \ell\left(\mathtt{A}(\mathcal{S}_n^{-i}), X_i\right)$, is a common alternative to the resubstitution estimate that aims to correct for this optimism. Here, $\mathcal{S}_n^{-i} = (X_1, \dots, X_{i-1}, X_{i+1}, \dots, X_n)$, i.e., it is the sequence $\mathcal{S}_n$ with example $X_i$ removed. Since $\mathbb{E}[\ell\left(\mathtt{A}(\mathcal{S}_n^{-i}), X_i\right)] = R_{n-1}(\mathtt{A}, \mathscr{P})$, then $\mathbb{E}[\widehat{R}_{\text{DEL}}(\mathtt{A}, \mathcal{S}_n)] = R_{n-1}(\mathtt{A}, \mathscr{P})$. When the latter is close to $R_n(\mathtt{A}, \mathscr{P})$, i.e., $\mathtt{A}$ is "stable", the deleted estimate may be a good alternative to the resubstitution estimate. However, due to the potentially strong correlations between elements of $(\ell(\mathtt{A}(\mathcal{S}_n^{-i}), X_i))_i$, the variance of the deleted estimate may be significantly higher than that of the resubstitution estimate (there is much redundancy in the information content of $\ell(\mathtt{A}(\mathcal{S}_n^{-i}), X_i)$ and $\ell(\mathtt{A}(\mathcal{S}_n^{-j}), X_j)$ for $i \neq j$). Another downside of the deleted estimate is its high computational cost that arises with a naive implementation, which is unavoidable in the lack of further structure. That is, to evaluate $\widehat{R}_{\text{DEL}}(\mathtt{A}, \mathcal{S}_n)$ for $\mathcal{S}_n$, one has to execute the learner $\mathtt{A}$ on $\mathcal{S}_n^{-i}$ to obtain hypothesis $\widehat{h}_i$, for $i = 1, \dots, n$; i.e. execute $\mathtt{A}$ for $n$ times. For large $n$, this is indeed prohibitive. Nevertheless, for some particular instances, the deleted estimate can be computed more efficiently, (i.e., in $O(\log n)$ time, in some cases). See for instance the work of Joulani et al. [2015] and the references therein.

The *$k$-fold cross validation* (KFCV) estimate provides a way of naturally interpolating between the resubstitution estimate and the deleted estimate [Stone, 1974, Geisser, 1975]. For simplicity, assume that the sequence $\mathcal{S}_n$ can be partitioned into $k$ equal folds $\mathcal{F}_{1,\dots,k} \doteq (\mathcal{F}_1 \ \dots \ \mathcal{F}_k)$, where each fold $\mathcal{F}_j$ is a sequence that has exactly $m$ examples from $\mathcal{S}_n$; i.e. $\mathcal{S}_n = (\mathcal{F}_1 \ \mathcal{F}_2 \ \cdots \ \mathcal{F}_k)$. In particular, we assume that $n = mk$. This assumption is merely made for convenience: all of our results extend to the general case with some extra effort. KFCV proceeds by learning $k$ hypotheses $\widehat{h}_1, \dots, \widehat{h}_k$, where $\widehat{h}_j = \mathtt{A}(\mathcal{S}_n^{-\mathcal{F}_j})$, and $\mathcal{S}_n^{-\mathcal{F}_j}$ is the sequence $(\mathcal{F}_1 \ \dots \ \mathcal{F}_{j-1} \ \mathcal{F}_{j+1} \ \dots \ \mathcal{F}_k)$; i.e. it is the sequence $\mathcal{S}_n$ with fold $\mathcal{F}_j$ removed. The empirical risk of $\widehat{h}_j$ is obtained by evaluating $\widehat{h}_j$ on $\mathcal{F}_j$ which was "held out" while running $\mathtt{A}$ on $\mathcal{S}_n^{-\mathcal{F}_j}$. The KFCV estimate for the risk is the average of the empirical risks of the $k$ hypotheses $\widehat{h}_1, \dots, \widehat{h}_k$:

$$\widehat{R}_{\text{CV}}(\mathtt{A}, \mathcal{F}_{1,\dots,k}) = \frac{1}{k}\sum_{j=1}^{k} \widehat{R}\left(\mathtt{A}(\mathcal{S}_n^{-\mathcal{F}_j}), \mathcal{F}_j\right) = \frac{1}{km}\sum_{j=1}^{k}\sum_{x \in \mathcal{F}_j} \ell\left(\mathtt{A}(\mathcal{S}_n^{-\mathcal{F}_j}), x\right) . \tag{1}$$

In the last expression of this display we are abusing the notation by using the membership operator '$\in$' with the sequence $\mathcal{F}_j$. In particular, in the sum every element of the set formed of the members of $\mathcal{F}_j$ appears with its multiplicity in $\mathcal{F}_j$.

Note that we obtain the deleted estimate as a special case of the KFCV estimate when $k = n$ and $m = 1$. The main goal of this paper is to develop a high probability upper bound on the absolute deviation $|\widehat{R}_{\text{CV}}(\mathtt{A}, \mathcal{F}_{1,\dots,k}) - R(\mathtt{A}(\mathcal{S}_n), \mathscr{P})|$ in terms of the "stability" of $\mathtt{A}$, which is defined next.

We note in passing that a similar bound can be derived for $|\widehat{R}_{\text{CV}}(\mathtt{A}, \mathcal{F}_{1,\dots,k}) - R_n(\mathtt{A}, \mathscr{P})|$. Our choice to bound the deviation of the KFCV estimate from the random risk $R(\mathtt{A}(\mathcal{S}_n), \mathscr{P})$ is a matter of practicality: It strikes us as somewhat impractical to estimate deviations from the expected risk, which can be markedly different than the actual risk of one's algorithm on a particular dataset. (However, the two quantities are also clearly quite closely related.)

## 3 Stability of Learning Rules

We start with the definition of $L_q$-stability, which specializes to Definition 1 by Celisse and Guedj [2016] when $m = 1$. For $m \in \mathbb{N}$, let $[m] \doteq \{1, \dots, m\}$. Fix $1 \leq m < n$, and let $\mathcal{S}_n^{-[m]}$ denote the sequence $\mathcal{S}_n$ after removing the first $m$ examples from it.[3]

[3] The notation $\mathcal{S}_n^{-i}$, $\mathcal{S}_n^{-\mathcal{F}_j}$, and $\mathcal{S}^{-[m]}$ might be overwhelming at first glance. Of course, these are all related; the "syntactic sugar" is introduced to simplify the notation. Besides these, we will only need $\mathcal{S}_n^{-\{\mathcal{F}_i, \mathcal{F}_j\}}$, which

**Definition 1** ($L_q$-stability Coefficient)**.** *Let $\mathcal{S}_n$ be a sequence of $n$ i.i.d random variables (RVs) drawn from $\mathcal{X}$ according to $\mathscr{P}$. Let $\mathtt{A}$ be a deterministic learning rule, and $\ell$ be a loss function as defined in Section 2. For $1 \leq m < n$, and $q \geq 1$, the $L_q$-stability coefficient of $\mathtt{A}$ with respect to $\ell$, $\mathscr{P}$, and $n, m$ is denoted by $\beta_q(\mathtt{A}, \ell, \mathscr{P}, n, m)$ and is defined as*

$$\beta_q^q(\mathtt{A}, \ell, \mathscr{P}, n, m) \;=\; \mathbb{E}\left[\left|\widehat{R}(\mathtt{A}(\mathcal{S}_n), \mathcal{F}') - \widehat{R}(\mathtt{A}(\mathcal{S}_n^{-[m]}), \mathcal{F}')\right|^q\right],$$

*where $\mathcal{F}' = (X'_1, \ldots, X'_m) \sim \mathscr{P}^m$ is independent of $\mathcal{S}_n$.*

Since the examples in $\mathcal{S}_n$ are *i.i.d*, the joint distribution of $(\mathtt{A}(\mathcal{S}_n), \mathtt{A}(\mathcal{S}_n^{-[m]}), X'_1, \ldots, X'_m)$ does not depend on which (fixed) $m$ examples are removed from $\mathcal{S}_n$, hence, for simplicity, in this definition we simply assume that it is always the first $m$ examples that are removed. Note that quite a few previous works restrict notions of algorithmic stability to learning rules that are permutation invariant, or *"symmetric"*; i.e. learning rules that yield identical output under different permutations of the examples presented to them [Rogers and Wagner, 1978, Devroye and Wagner, 1979, Kearns and Ron, 1999, Bousquet and Elisseeff, 2002, Shalev-Shwartz et al., 2010]. For the same reason of why it does not matter which examples are removed, it does not matter whether the learning rule is symmetric or not.

Since often $\mathtt{A}$, $\ell$, $\mathscr{P}$ are fixed, we will drop them from the notation and will just use $\beta_q^q(n, m)$. However, this should not be mistaken to taking a supremum over any subset of the dropped quantities: The stability coefficients are meant to be algorithm, loss and distribution dependent. By avoiding a worst-case approach in the definitions, we will be able to get a finer picture than if we took a worst-case approach.

The $L_q$-stability coefficient quantifies the variation of the random risk of $\mathtt{A}$ induced by removing $m$ samples from the training set. Often, this is known as a *removal type* notion of stability which is different from (but related to) the *replacement type* notion of stability where the example $X_i$ is replaced with the example $X'_i$ s.t. $X'_i \sim \mathscr{P}$ and $X'_i$ is independent of $\mathcal{S}_n$. This definition of stability is therefore in accordance with previous notions of stability [Rogers and Wagner, 1978, Devroye and Wagner, 1979, Kearns and Ron, 1999, Bousquet and Elisseeff, 2002]. For instance, if $q = 1$ and $m = 1$, this is the $L_1$-stability or hypothesis stability introduced by Rogers and Wagner [1978] and Devroye and Wagner [1979] to derive PAC polynomial generalization bounds for the $k$-nearest neighbor rule and potential functions rules, respectively. The difference between $L_q$-stability and earlier notions of stability, is that $L_q$-stability is in terms of the higher order moments of the RV $|\widehat{R}(\mathtt{A}(\mathcal{S}_n), \mathcal{F}') - \widehat{R}(\mathtt{A}(\mathcal{S}_n^{-[m]}), \mathcal{F}')|$. The reason we care about higher moments is because we are interested in controlling the tail behavior of the KFCV estimate. It is then quite expected that the tail behavior of the KFCV estimate is also dependent on the tail behavior of RVs characterizing stability. As is well-known, knowledge of the higher moments of a RV is equivalent to knowledge of the tail behavior of the RV.

**$q$-Norm of RVs:** In the sequel, we will heavily rely on the $q$-norm for RVs. For a real RV $X$, and for $1 \leq q \leq +\infty$, the $q$-norm of $X$ is defined as: $\|X\|_q \doteq (\mathbb{E}[|X|^q])^{1/q}$, and $\|X\|_\infty$ is the essential supremum of $|X|$. Note that these norms satisfy $\|\cdot\|_q \leq \|\cdot\|_p$ for $1 \leq q \leq p \leq +\infty$.

As a result of the last inequality, for a fixed $q$, $\ell$, $\mathtt{A}$ and $\mathscr{P}$, $\beta_q(n, m) \doteq \beta_q(\mathtt{A}, \ell, \mathscr{P}, n, m)$ is an increasing function of $q$. Furthermore, we also expect that $\beta_q(n, m)$ will be a decreasing function of $n$ and an increasing function of $m$. The definition of $\beta_q(n, m)$ and triangle inequalities (which hold for $\|\cdot\|_q$) immediately give that $\beta_q(n, m) \leq \sum_{j=0}^{m-1} \beta_q(n - j, 1)$, which results $\beta_q(n, m) \leq m\beta_q(n - m, 1)$ when $n \mapsto \beta_q(n, m)$ is decreasing. However, one can check that this latter inequality is too crude. In particular, for the problem of estimating the mean under the squared loss, $\beta_q^2(n - m, m) \sim \|X\|_{2q}^2 m/n^2$ (as hinted about in the introduction). A similar relation is expected to hold for penalized estimation with appropriate penalties.

will denote that both folds indexed by $i \neq j$ are removed from $\mathcal{S}_n$. Other than these, no further notation will be introduced in this regard.

## 4 Main Tool

The main tool for our work is an extension of the Efron-Stein inequality [Efron and Stein, 1981, Steele, 1986], to a stronger version known as the exponential Efron-Stein inequality [Boucheron et al., 2003]. The Efron-Stein inequality is a strong tool itself to bound the variance $\mathbb{V}[Z] \doteq \mathbb{E}[(Z - \mathbb{E}Z)^2]$ of a random variable $Z$ which is a function (call this $f$) of a number of independent RVs. The idea of the Efron-Stein inequality is to "decompose" the variance into the sum $V$ of variance-like terms that express the sensitivity of the function $f$ to its individual variables in an appropriate manner. Oftentimes, these individual sensitivities are easier to control than the variance directly. The crucial feature of the inequality is that it avoids pessimistic worst-case bounds like those that underly McDiarmid's inequality [McDiarmid, 1989]. While bounding the variance itself is crucial, we will need exponential concentration bounds on the tails of $Z$. Such bounds were derived in the work of Boucheron et al. [2003, 2013]. Here, based on the technique developed in this groundbreaking work we derive a new tail inequality which will better suit our purposes.

We start by introducing the Efron-Stein inequality and some variations. The inequalities shown here will be useful in our derivations on their own. Let $f : \mathcal{X}^n \longmapsto \mathbb{R}$ be a real-valued function of $n$ variables, where $\mathcal{X}$ is a measurable space (not necessarily the same as in the previous section). Let $X_1, \ldots, X_n$ be independent (not necessarily identically distributed) RVs taking values in $\mathcal{X}$ and define the RV $Z = f(X_1, \ldots, X_n) \equiv f(\mathcal{S}_n)$. Define the shorthand for the conditional expectation $\mathbb{E}_{-i}Z \doteq \mathbb{E}\left[Z|\mathcal{S}_n^{-i}\right]$, where $\mathcal{S}_n^{-i}$ is defined as in the previous section. Informally, $\mathbb{E}_{-i}Z$ "integrates" $Z$ over $X_i$ and *also over any other source of randomness* in $Z$ except $\mathcal{S}_n^{-i}$. The celebrated Efron-Stein inequality bounds the variance of $Z$ as shown in the following theorem:

**Theorem 1** (Efron-Stein Inequality)**.** *Let $V = \sum_{i=1}^n (Z - \mathbb{E}_{-i}Z)^2$. Under the setting described in this section, it holds that $\mathbb{V}[Z] \leq \mathbb{E}V$.*

The proof of Theorem 1 can be found, e.g., in the paper of Boucheron et al. [2004]. Another variant of the Efron-Stein inequality which will turn out to be more useful for our context, is concerned with the removal of one example from $\mathcal{S}_n$. To state the result, let $f_i : \mathcal{X}^{n-1} \longmapsto \mathbb{R}$, for $1 \leq i \leq n$, be an arbitrary measurable function, and define the RV $Z_{-i} = f_i(\mathcal{S}_n^{-i})$. Then, the Efron-Stein inequality can be also stated in the following interesting form:

**Corollary 1** (Efron-Stein Inequality – Removal Case. Theorem 6 from [Boucheron et al., 2004])**.** *Assume that $\mathbb{E}_{-i}[Z_{-i}]$ exists for all $1 \leq i \leq n$, and let $V_{DEL} = \sum_{i=1}^n (Z - Z_{-i})^2$. Then it holds that*

$$\mathbb{V}[Z] \;\leq\; \mathbb{E}V \;\leq\; \mathbb{E}V_{DEL}\,. \tag{2}$$

It may look surprising at a first sight that $\mathbb{V}[Z]$ can be bounded in terms of $V_{\text{DEL}}$ which relies on arbitrary functions $f_i$ *unrelated* to $f$. However, the inequality above should already be quite revealing as it suggests that the best choice for $f_i$ is $f_i(\mathcal{S}_n^{-i}) = \mathbb{E}_{-i}Z$. Note the resemblance to $\mathbb{V}[Z] = \mathbb{E}[(Z - \mathbb{E}Z)^2] \leq \mathbb{E}[(Z - a)^2]$, which holds for any real number $a$. The simple proof given by Boucheron et al. [2004] reveals that this parallel is not a coincidence.

### 4.1 An Exponential Efron-Stein Inequality

The work of Boucheron et al. [2003] has focused on controlling the tail of general functions of independent RVs in terms of the tail behavior of the Efron-Stein variance-like terms such as $V$ and $V_{\text{DEL}}$, as well as other terms known as $V^+$ and $V^-$. The variance-like terms $V$, $V^+$ and $V^-$ measure the sensitivity of a function of $n$ independent RVs w.r.t the *replacement* of one RV from the $n$ independent RVs. The term $V_{\text{DEL}}$ on the other hand, measures the sensitivity of a function of $n$ independent RVs w.r.t the *removal* of one RV from the $n$ independent RVs. In this work, we favor $V_{\text{DEL}}$ over the other terms since it is more suitable for our choice of stability coefficient, the $L_q$-stability, which is also a removal version. The removal version of stability is preferred as it is more natural in the learning context where one is given a fixed sample. In particular, the removal version seems to be a better fit when it comes to empirically estimating stability (which is an interesting future direction), where to work with the replacement version one would need extra data, or extra assumptions.

The tail of a RV is often controlled through bounding the logarithm of the moment generating function (MGF) of the RV. This is known as the *cumulant generating function* (CGF) of the RV and is defined as: $\psi_Z(\lambda) \doteq \log \mathbb{E}\left[\exp(\lambda Z)\right]$, where $\lambda \in \text{dom}(\psi_Z) \subset \mathbb{R}$, and belongs to a suitable neighborhood of zero. The main result of Boucheron et al. [2003] bounds $\psi_Z$ in terms of the MGF for $V$, $V^+$ and $V^-$, but not in terms of the MGF for $V_{\text{DEL}}$. Since we are particularly interested in the RV $V_{\text{DEL}}$, the following theorem bounds the tail of $\psi_Z$ in terms of the MGF for $V_{\text{DEL}}$. The proof, which mimics the proof technique of Boucheron et al. [2003], is given in the Appendix.

**Theorem 2.** *Let $Z$, $V_{DEL}$ be as in Corollary 1 and assume that $|Z - Z_{-i}| \leq 1$ almost surely for all $i$. For all $\theta > 0$, such that $\lambda \in (0, 1]$, $\theta\lambda < 1$, and $\mathbb{E}e^{\lambda V_{DEL}} < \infty$, the following holds:*

$$\log \mathbb{E}\left[\exp\left(\lambda(Z - \mathbb{E}Z)\right)\right] \leq \lambda\theta(1-\lambda\theta)^{-1} \log \mathbb{E}\left[\exp\left(\lambda\theta^{-1}V_{DEL}\right)\right] . \tag{3}$$

Theorem 2 states that the CGF of the centered RV $Z - \mathbb{E}Z$ is upper bounded by the CGF of the RV $V_{\text{DEL}}$. Hence, when $V_{\text{DEL}}$ behaves "nicely", the tail of $Z$ can be controlled. The value of $\theta$ in the upper bound is a free parameter that can be optimized.

For Theorem 2 to be useful in our context, further control is required to upper bound the tail of $V_{\text{DEL}}$. Our approach to control the tail of $V_{\text{DEL}}$ will be, again, through its CGF. In particular, we aim to show that when $V_{\text{DEL}}$ is a sub-gamma RV (defined shortly) we can obtain a high probability tail bound on the deviation of the RV $Z$. The obtained tail bound will be instrumental in deriving the exponential tail bound for the KFCV estimate.

**Sub-Gamma RVs:** We follow here the notation of the work of Boucheron et al. [2013]. A real valued centered RV $X$ is said to be *sub-gamma* on the right tail with variance factor $v$ and scale parameter $c$ if for every $\lambda$ such that $0 < \lambda < 1/c$, the following holds

$$\psi_X(\lambda) \leq \tfrac{1}{2}\lambda^2 v(1 - c\lambda)^{-1} . \tag{4}$$

This is denoted by $X \in \Gamma_+(v, c)$. Similarly, $X$ is said to be a sub-gamma RV on the left tail with variance factor $v$ and scale parameter $c$ if $-X \in \Gamma_+(v, c)$. This is denoted as $X \in \Gamma_-(v, c)$. Finally, $X$ is simply a sub-gamma RV with variance factor $v$ and scale parameter $c$ if both $X \in \Gamma_+(v, c)$ and $X \in \Gamma_-(v, c)$. This is denoted by $X \in \Gamma(v, c)$. The sub-gamma property can be equivalently defined in terms of tail bounds or moment conditions as follows from Theorem 2.3 from the book of Boucheron et al. [2013]:

**Theorem 3.** *Let $X$ be a centered RV. If for some $v > 0$ and $c \geq 0$*

$$\mathbb{P}\left[X > \sqrt{2vt} + ct\right] \vee \mathbb{P}\left[-X > \sqrt{2vt} + ct\right] \leq e^{-t} \text{ , for every } t > 0 \text{ ,} \tag{5}$$

*then for every integer $q \geq 1$*

$$\|X\|_{2q} \;\leq\; (q!A^q + (2q)!B^{2q})^{1/2q} \;\leq\; \sqrt{16.8qv} \vee 9.6qc \;\leq\; 10(\sqrt{qv} \vee qc) \,.$$

*where $A = 8v$, $B = 4c$. Conversely, if for some positive constants $u$ and $w$, for any integer $q \geq 1$,*

$$\|X\|_{2q} \;\leq\; \sqrt{qu} \vee qw \;,$$

*then (5) holds with $v = 4(1.1u + 0.73^2w^2)$ and $c = 1.46w$.*

The reader may notice that Theorem 3 is slightly different than the version in the book of Boucheron et al. [2013]. Our extension is based on simple (and standard) calculations that are merely for convenience with respect to our purpose.

#### 4.1.1 An Exponential Tail Bound for $Z$

In this section we assume that $V_{\text{DEL}} - \mathbb{E}V_{\text{DEL}}$ is a sub-gamma RV with variance factor $v > 0$, scale parameter $c \geq 0$, and $c\lambda < 1$. Hence, from (4) it holds that

$$\psi_{V_{\text{DEL}} - \mathbb{E}V_{\text{DEL}}}(\lambda) = \log \mathbb{E}\left[\exp(\lambda(V_{\text{DEL}} - \mathbb{E}V_{\text{DEL}}))\right] \leq \tfrac{1}{2}\lambda^2 v(1 - c\lambda)^{-1} .$$

The sub-gamma property of $V_{\text{DEL}}$ provides the desired control on its tail. That is, after arranging the terms of the above inequality, the CGF of $V_{\text{DEL}}$ which controls the tail of $V_{\text{DEL}}$, is upper bounded by the deterministic quantities: $\mathbb{E}V_{\text{DEL}}$, the variance $v$, and the scale parameter $c$. Therefore, it is possible now to use the sub-gamma property of $V_{\text{DEL}}$ in the result of the exponential Efron-Stein inequality in Theorem 2. In particular, the following lemma gives an exponential tail bound on the deviation of a function of independent RVs, i.e. $Z = f(X_1, \ldots, X_n)$, in terms of $\mathbb{E}V_{\text{DEL}}$, the variance factor $v$, and the scale parameter $c$. This lemma will be our main tool to derive the exponential tail bound on the KFCV estimate. The proof is given in the Appendix.

**Lemma 1.** *Let $Z$, $Z_{-i}$, $V_{DEL}$ be as in Theorem 2. If $V_{DEL} - \mathbb{E}V_{DEL}$ is a sub-gamma RV with variance parameter $v > 0$ and scale parameter $c \geq 0$, then for any $\delta \in (0,1)$, $a > 0$, with probability $1 - \delta$,*

$$|Z - \mathbb{E}Z| \;\leq\; \tfrac{4}{3}(ac + 1/a)\log\left(\tfrac{2}{\delta}\right) + 2\sqrt{(\mathbb{E}V_{DEL} + a^2 v/2)\log\left(\tfrac{2}{\delta}\right)}\;.$$

Parameter $a$ in the upper bound is a free parameter that can be optimized to provide the tightest possible bound. A typical choice of $a$ would be the inverse standard deviation of $Z$.

# 5 An Exponential Tail Bound for KFCV

The main goal of this paper is to develop a high probability upper bound on the absolute deviation of $\widehat{R}_{\text{CV}}\left(\mathtt{A}, \mathcal{F}_{1,\ldots,k}\right)$ from the risk $R(\mathtt{A}(\mathcal{S}_n), \mathscr{P}) = \mathbb{E}[\ell\left(\mathtt{A}(\mathcal{S}_n), X\right)|\mathcal{S}_n]$ using the tools developed earlier. To do so, we decompose $\Delta \doteq |\widehat{R}_{\text{CV}}\left(\mathtt{A}, \mathcal{F}_{1,\ldots,k}\right) - R(\mathtt{A}(\mathcal{S}_n), \mathscr{P})|$ into three terms

$$|\widehat{R}_{\text{CV}}\left(\mathtt{A}, \mathcal{F}_{1,\ldots,k}\right) - R(\mathtt{A}(\mathcal{S}_n), \mathscr{P})| \leq \text{I} + \text{II} + \text{III}\,, \tag{6}$$

where

$$\begin{aligned}
\text{I} &= |\mathbb{E}\widehat{R}_{\text{CV}}(\mathtt{A}, \mathcal{F}_{1,\ldots,k}) - \widehat{R}_{\text{CV}}(\mathtt{A}, \mathcal{F}_{1,\ldots,k})|, \\
\text{II} &= |R(\mathtt{A}(\mathcal{S}_n), \mathscr{P}) - \mathbb{E}R(\mathtt{A}(\mathcal{S}_n), \mathscr{P})|, \quad \text{and} \\
\text{III} &= |\mathbb{E}R(\mathtt{A}(\mathcal{S}_n), \mathscr{P}) - \mathbb{E}\widehat{R}_{\text{CV}}(\mathtt{A}, \mathcal{F}_{1,\ldots,k})|.
\end{aligned}$$

As such, if the three terms in the RHS of (6) are properly upper bounded, we will have the desired final upper bound. Terms I and II shall be bounded using the exponential Efron-Stein inequality introduced in the previous section. Further, we hope that the final upper bounds can be in terms of the $L_q$-stability coefficient of $\mathtt{A}$. Term III, however, is non-random and thus shall be directly bounded using some $L_q$-stability coefficient.

For terms I and II, the key quantity for using the exponential Efron-Stein inequality in Lemma 1 is the RV $V_{\text{DEL}}$. In particular, the requirement for using $V_{\text{DEL}}$ is two-fold. First, since $V_{\text{DEL}} = \sum_{i=1}^{n}(Z - Z_{-i})^2$, where $Z_{-i} = f_i(\mathcal{S}_n^{-i})$ for some function $f_i$, we need to choose $f_i$ appropriately. Second, once $Z_{-i}$ is defined, to be able to use Lemma 1 we need to show that $V_{\text{DEL}}$ is a sub-gamma RV. For this, from Theorem 3 we know that it suffices to show that for all integers $q \geq 1$,

$$\|V_{\text{DEL}}\|_{2q} \leq \sqrt{qu} \vee qw\,, \tag{7}$$

for some positive constants $u$ and $w$. Here, we will relate $\|V_{\text{DEL}}\|_{2q}$ to $L^q$-stability coefficients and then we "reverse engineer" appropriate assumptions on the $L^q$-stability coefficients that imply (7).

## 5.1 Deriving The Upper Bounds for Terms I, II, and III

In this section we derive the desired upper bounds for Terms I, II, and III. Unless otherwise stated, all proofs for the results in this section can be found in the Appendix. First, we consider Term I in the RHS of inequality (6). This is the deviation $|\mathbb{E}\widehat{R}_{\text{CV}}(\mathtt{A}, \mathcal{F}_{1,\ldots,k}) - \widehat{R}_{\text{CV}}(\mathtt{A}, \mathcal{F}_{1,\ldots,k})|$. Note that $\widehat{R}_{\text{CV}}\left(\mathtt{A}, \mathcal{F}_{1,\ldots,k}\right) \equiv \widehat{R}_{\text{CV}}\left(\mathtt{A}, \mathcal{F}_1, \ldots, \mathcal{F}_k\right)$ is a function of $k$ independent random sequences. Hence, the exponential Efron-Stein inequality in Lemma 1 seems applicable. In particular, we propose to use it with

$$Z \;=\; \widehat{R}_{\text{CV}}\left(\mathtt{A}, \mathcal{F}_{1,\ldots,k}\right)\,, \qquad\qquad Z_{-i} \;=\; \frac{1}{k-1}\sum_{j=1,\; j\neq i}^{k} \widehat{R}\left(\mathtt{A}(\mathcal{S}_n^{-\{\mathcal{F}_i, \mathcal{F}_j\}}), \mathcal{F}_j\right)\,, \tag{8}$$

where $-\{\mathcal{F}_i, \mathcal{F}_j\}$ indicates the removal of folds $\mathcal{F}_i$ and $\mathcal{F}_j$ from $\mathcal{S}_n = (\mathcal{F}_1\ \mathcal{F}_2\ \cdots\ \mathcal{F}_k)$ and in particular $\mathcal{S}_n^{-\{\mathcal{F}_i, \mathcal{F}_j\}}$ indicates the same sequence regardless of whether $i < j$ or $j < i$.

Recall that $V_{\text{DEL}} = \sum_i (Z - Z_{-i})^2$. According to the plan outlined above, let us now turn to upper bounding the moments of $V_{\text{DEL}}$. In this regard, we have the following result:

**Lemma 2.** *Let $Z$, $Z_{-i}$ be defined as in (8), and let $V_{DEL} = \sum_{i=1}^{k}\left(Z - Z_{-i}\right)^2$. Then for any real $q \geq 1/2$, $k \geq 1$, and $n > m \geq 1$, the following holds:*

$$\|V_{DEL}\|_{2q} \;\leq\; k\beta_{4q}^2(n-m, m)\,. \tag{9}$$

To use Lemma 1 we also need a bound on $\mathbb{E}V_{\text{DEL}}$. However, this can be directly obtained from (9) since $V_{\text{DEL}} \geq 0$ and thus $\|V_{\text{DEL}}\|_{2q} = \mathbb{E}V_{\text{DEL}}$ when $q = 1/2$:

$$\mathbb{E}V_{\text{DEL}} \;\leq\; k\beta_2^2(n-m,m)\;. \tag{10}$$

From Lemma 2 and Theorem 3 it follows immediately that $V_{\text{DEL}}$ is a sub-gamma RV provided that $(k\beta_{4q}^2(n-m,m))_{q\geq 1}$ is well-behaved as postulated in our next assumption:

**Assumption 1.** *$\exists\, u_1, w_1 \geq 0$ s.t. for any integer $q \geq 1$, it holds that $k\beta_{4q}^2(n-m,m) \leq \sqrt{qu_1} \vee qw_1$.*

This assumption is needed since our results are in terms of the stability of a *generic* learning rule $\mathtt{A}$ with minimal knowledge about it and about its stability. We expect the above to hold for reasonable algorithms. For example, Celisse and Guedj [2016] in their Theorem 1 prove that for ridge regression with a fixed constant ridge parameter and bounded covariates, $\beta_q(n-1,1) \leq C\,\|Y\|_{2q}^2/n$, where $Y$ is the response variable and $C$ depends on the bound on the covariates and the ridge parameter. We conjecture that $\beta_q(n-m,m)$, as a function of $n, m$ behaves as $\sqrt{m}/n$, but that it behaves identically to $\beta_q(n-1,1)$ otherwise. (In particular, this is so in the case of estimating the mean.) As a result, we see that Assumption 1 will be satisfied as long as $Y$ is a sub-gamma RV.

**Corollary 2.** *Using the previous definitions, and under Assumption 1, $V_{DEL} \in \Gamma(v_1, c_1)$, where $v_1 = 4(1.1u_1 + 0.73^2 w_1^2)$ and $c_1 = 1.46w_1$.*

The statement of Corollary 2 follows from Lemma 2, Assumption 1 and Theorem 3. Now, plugging the result of Corollary 2 and inequality (10) into Lemma 1 gives the desired final upper bound for Term I. Since Lemma 1 requires that $|Z - Z_{-i}| \leq 1$ almost surely, this result uses the boundedness of the loss, which we assumed. (Since this is the only time boundedness is used, it appears that boundedness can be removed at the expense of slightly tweaking our definitions of stability, as we shall discuss it later.)

**Lemma 3.** *Under Assumption 1, and for $k \geq 1$, and $n > m \geq 1$, let $r_1 = k\beta_2^2(n-m,m)$. Then for any $\delta \in (0,1)$ and $a > 0$, with probability $1-\delta$ the following holds*

$$\begin{aligned}\left|\mathbb{E}\widehat{R}_{CV}(\mathtt{A}, \mathcal{F}_{1,\ldots,k}) - \widehat{R}_{CV}(\mathtt{A}, \mathcal{F}_{1,\ldots,k})\right| \;\leq\; & \tfrac{4}{3}(1.46aw_1 + \tfrac{1}{a})\log\left(\tfrac{2}{\delta}\right) \\ & + 2\sqrt{(r_1 + 2.2a^2u_1 + 1.07(aw_1)^2)\log\left(\tfrac{2}{\delta}\right)}\;.\end{aligned}$$

Let us now discuss the choice of $a$ to get some insight about the qualitative behavior of this bound in the context of how it may scale with $n$. Note that from Assumption 1 we can see that $u_1$ and $w_1$ are controlled by $k\beta_{4q}^2(n-m,m) = k\beta_{4q}^2(n-n/k, n/k)$. In particular, for $k$ fixed, they depend on $n$. If, for example $k\beta_{4q}^2(n-n/k,n/k) \sim k^{-1}n^{-p}$ with some $p > 0$ (e.g., for ridge regression, we conjecture $p = 1$) then $u_1 \sim k^{-2}n^{-2p}$ and $r_1, w_1 \sim k^{-1}n^{-p}$. The terms that depend on $a$ from the bound then scale as $ak^{-1}n^{-p} + \frac{1}{a}$ with $n$ and $k$. Hence, choosing $a = k^{1/2}n^{p/2}$ makes both the $a$-dependent part, as well as the whole of the bound scale with $k^{-1/2}n^{-p/2}$ as a function of $n$ and $k$. In general, we expect $w_1 \approx \sqrt{u_1}$, in which case choosing $a = w_1^{-1/2}$ makes the bound scale with $w_1^{1/2}$. Then, $w_1^{1/2} = o(1)$ (as $n \to \infty$) will be sufficient for consistency, translating to $\beta_{4q}(n-n/k, n/k) = o(1)$ with $k$ fixed. The case when $k$ changes with $n$, e.g., when $k_n$-fold CV is used with a sample size of $n$, we get $k_n^{1/2}\beta_{4q}(n-n/k_n, n/k_n) = o(1)$. For example, when $k_n = n$ (corresponding to the deleted estimate), this translates to $\beta_{4q}(n-1,1) = o(1/n^{1/2})$, a condition that has been identified as sufficient for consistency of deleted estimates earlier (e.g., [Celisse and Guedj, 2016]).

Let us now turn to bounding Term II in inequality (6). This is the deviation $|\mathbb{E}R\left(\mathtt{A}(\mathcal{S}_n), \mathscr{P}\right) - R\left(\mathtt{A}(\mathcal{S}_n), \mathscr{P}\right)|$. Note that $R\left(\mathtt{A}(\mathcal{S}_n), \mathscr{P}\right)$ is a function of $n$ independent RVs, and therefore, Lemma 1 will be our tool to bound this deviation. Similar to the steps for deriving the bound for Term I, we need to define the RVs $Z$ and $Z_{-i}$, and show that $V_{\text{DEL}}$ is a sub-gamma RV. In this case we let $Z$ and $Z_{-i}$ be defined as follows:

$$Z \;=\; R\left(\mathtt{A}(\mathcal{S}_n), \mathscr{P}\right)\,, \qquad\qquad Z_{-i} \;=\; R\left(\mathtt{A}(\mathcal{S}_n^{-i}), \mathscr{P}\right)\,. \tag{11}$$

Similarly to Lemma 2 we have the following result:

**Lemma 4.** *Let $Z$ and $Z_{-i}$ be defined as in (11) and let $V_{DEL} = \sum_{i=1}^{n}(Z - Z_{-i})^2$. Then for any real $q \geq 1/2$, and $n \geq 2$, the following holds:*

$$\|V_{DEL}\|_{2q} \;\leq\; n\beta_{4q}^2(n,1) \;. \tag{12}$$

Lemma 1 also requires a bound on $\mathbb{E}V_{\text{DEL}}$. As before, we obtain this from (12) directly by noticing that $V_{\text{DEL}} \geq 0$, and for $q = 1/2$, $\|V_{\text{DEL}}\|_{2q} = \mathbb{E}V_{\text{DEL}}$:

$$\mathbb{E}V_{\text{DEL}} \;\leq\; n\beta_2^2(n,1) \;. \tag{13}$$

From Lemma 4 and Theorem 3, it follows that $V_{\text{DEL}}$ is a sub-gamma RV if $(n\beta_{4q}^2(n,1))_{q\geq 1}$ is well-behaved as postulated in our next assumption:

**Assumption 2.** $\exists\, u_2, w_2 \geq 0$ *s.t. for any integer $q \geq 1$, it holds that $n\beta_{4q}^2(n,1) \leq \sqrt{qu_2} \vee qw_2$.*

Note by our earlier remark, for ridge regression, $n\beta_{4q}^2(n,1) \leq C^2 \|Y\|_{8q}^4 / n$. Again, we see that Assumption 2 will be satisfied as long as $Y$ is a sub-gamma RV. Note that if $m \mapsto \beta_q(n-m,m)$ is increasing with $m$ and $n \mapsto \beta_q(n-1,1)$ is decreasing (we expect these to hold for reasonable algorithms) then $\beta_{4q}^2(n,1) \leq \beta_{4q}^2(n-1,1) \leq \beta_{4q}^2(n-m,m)$, and thus Assumption 2 will be implied by Assumption 1.

From Lemma 4, Assumption 2 and Theorem 3, we can state the following corollary which gives the parameters of the sub-gamma RV $V_{\text{DEL}}$.

**Corollary 3.** *Using the previous definitions, and under Assumption 2, $V_{DEL} \in \Gamma(v_2, c_2)$, where $v_2 = 4(1.1u_2 + 0.73^2 w_2)$ and $c_2 = 1.46w_2$.*

The final upper bound for Term II is given by the following lemma which simply plugs in the results of Corollary 3 and inequality (13) into Lemma 1.

**Lemma 5.** *Under Assumption 2, and for $n \geq 2$, let $r_2 = n\beta_2^2(n,1)$. Then for $\delta \in (0,1)$ and $a > 0$, with probability $1-\delta$ the following holds*

$$\begin{aligned}|\mathbb{E}R(\mathtt{A}(\mathcal{S}_n),\mathscr{P}) - R(\mathtt{A}(\mathcal{S}_n),\mathscr{P})| \;\leq\; & \tfrac{4}{3}(1.46aw_2 + 1/a)\log\left(\tfrac{2}{\delta}\right) \\ & + 2\sqrt{(r_2 + 2.2a^2u_2 + 1.07a^2w_2^2)\log\left(\tfrac{2}{\delta}\right)} \;.\end{aligned}$$

Concerning the choice of $a$, the discussion after Lemma 3 applies. However, for example, in the case of ridge regression, we see that $w_2, u_2^{1/2} \sim 1/n$ (while we had $w_1, u_1^{1/2} \sim 1/n^2$). As a result, in this case we would choose $a = w_2^{-1/2} = n^{1/2}$, which gives that the bound is of order $O(n^{-1/2})$. This is similar (of course) to the case discussed after Lemma 3 when $k = k_n = n$. For consistency, assuming $w_2 \approx u_2^{1/2}$ and choosing $a = w_2^{-1/2}$, we require $w_2^{1/2} = o(1)$, which translates into $\beta_{4q}(n,1) = o(1/n^{1/2})$, which was the condition that was identified as sufficient for the in-probability convergence of the term bounded in Lemma 3 to zero as $n \to \infty$ when $k = k_n = n$. It appears that controlling Term II puts more demand on the stability of an algorithm than controlling Term I.

For Term III in inequality (6) there are no random quantities to account for since both terms in the absolute value are expectations of RVs. Hence, an upper bound on this deviation will always hold.

**Lemma 6.** *Using the previous setup and definitions, let $\mathtt{A}$ be a learning rule with $L_2$ stability coefficient $\beta_2(n,m)$. Then, for $k \geq 1$, and $n > m \geq 1$, the following holds*

$$\left|\mathbb{E}R(\mathtt{A}(\mathcal{S}_n),\mathscr{P}) - \mathbb{E}\widehat{R}_{CV}(\mathtt{A},\mathcal{F}_{1,\ldots,k})\right| \;\leq\; \beta_2(n,m) \;.$$

### 5.2 Main Result

**Theorem 4.** *Let $\mathcal{X}$, $\mathcal{H}$ and $\ell$ be as previously defined. Let $\mathcal{S}_n = (\mathcal{F}_1,\ldots,\mathcal{F}_k)$ be the dataset defined in Section 2.2, where $k \geq 1$, $n > m \geq 1$, and $n = km$. Let $\widehat{R}_{CV}(\mathtt{A},\mathcal{F}_{1,\ldots,k})$ be the cross validation estimate defined in (1), and $R(\mathtt{A}(\mathcal{S}_n),\mathscr{P})$ be the risk for hypothesis $\mathtt{A}(\mathcal{S}_n)$. Then, under Assumption 1 and 2, for all $\delta \in (0,1)$ and $a > 0$, with probability $1-\delta$, the following holds*

$$\left|\widehat{R}_{CV}(\mathtt{A},\mathcal{S}_n) - R(\mathtt{A}(\mathcal{S}_n),\mathscr{P})\right| \leq 2(aw_1 + aw_2 + 2/a)\log\left(\tfrac{4}{\delta}\right) + \beta_2(n,m) + (\pi_1 + \pi_2)\sqrt{\log\left(\tfrac{4}{\delta}\right)}\;,$$

*where*

$$\pi_1 = 2\sqrt{k\beta_2^2(n-m,m) + 2.2a^2u_1 + 1.07a^2w_1^2}\,, \quad \textit{and}$$
$$\pi_2 = 2\sqrt{n\beta_2^2(n,1) + 2.2a^2u_2 + 1.07a^2w_2^2}\,.$$

Concerning the choice of $a$, the discussions after Lemmas 3 and 5 apply. For consistency, under assumptions stated there, we need $\beta_{4q}(n,1) = o(1/n^{1/2})$ and $\beta_{4q}(n-n/k, n/k) = o(1)$ with $k$ fixed, while we need $\beta_{4q}(n,1) = o(1/n^{1/2})$ only when $k = k_n = n$ (deleted estimate).

The proof of Theorem 4 simply plugs in the results of Lemma 3, Lemma 5, and Lemma 6 into inequality (6). The final bound has four terms. The first term is due to the higher order moments of the $L_q$-stability, in particular $\beta_{4q}^2$ which depends on the tail behavior of the RV $|\ell(\mathtt{A}(\mathcal{S}_n), X) - \ell(\mathtt{A}(\mathcal{S}_n^{-1}), X)|^{4q}$. Note that from Theorem 3, $w_1$ and $w_2$ are in fact controlled by the stability coefficients $(\beta_{4q}^2)_q$. Therefore, as the stability is improving, $w_1$ and $w_2$ will be small, thereby making the bound tighter. Note that the same applies to $u_1$ and $u_2$ in $\pi_1$ and $\pi_2$, respectively.

## 6 Concluding Remarks and Outlook

As stated earlier, our concentration bound for the KFCV estimate shows that in order for the estimate to concentrate around the true risk, the learning rule $\mathtt{A}$ has to be stable. Depending on the degree of stability, one may then need to increase $k$ as a function of the sample size. For "highly stable" learning rules, the number of folds $k$ matters only up to a constant factor at most. More precisely, a lengthy and tedious calculation shows that for the sample mean, the tail bound that can be derived from our main result, at confidence level $\delta \in (0,1)$, is (up to universal constant factors) $\frac{1}{nk}\sqrt{\log(1/\delta)} + \frac{1}{n}\log(1/\delta)$. Thus, $k$ at best modifies a lower-order term (lower order in terms of dependence on the confidence parameter $\delta$): In particular, no matter the value of $k$, the length of the confidence interval is within a factor of 2 of the length that can be obtained with $k = 2$. Of course, this is unsurprising as even the resubstitution estimate is a reasonable estimate of the risk in the case of sample mean estimation.

The situation is more interesting only in problems where "overfitting" is a real possibility. However, when overfitting is a possibility then the learning rule is expected to be unstable, hence KFCV may be unreliable (leading to a wide confidence bound), while when KFCV is reliable, overfitting may not occur. We expect that for methods like ridge regression where stability can be increased or decreased, the choice of $k$ may have a larger effect as stability varies. In particular, we expect that in the near-unstable regime, the resubstitution estimate will be poor, and KFCV should then be used with a larger value of $k$, while in the highly stable regime (like in the case of sample-mean estimation), the choice of $k$ will be immaterial. In particular, a small value may be used, or even the resubsitution estimate may then be a good choice. In any case, it remains an interesting problem to quantify this effect in some specific cases, such as ridge regression.

According to the above discussion, and now considering the practical side of our profession, one has to question the widely used practice of setting $k$ to a predefined value to report the empirical generalization error for a learning rule, regardless of the learning rule, the data distribution, or the sample size $n$; for instance setting $k = 10$ or $k = 5$. Note that for a fixed $k$, as the sample size is increasing, the size of each fold, $m = n/k$ is also increasing. To get reasonable risk estimates the learning rule, in terms of the hypothesis loss, will then need to be stable w.r.t the removal of $m$, i.e., a large fraction of examples from the training set. If there is no justification for this stability assumption, one should wonder the faithfulness and the reliability of such empirical results. This practice is even more alarming in the absence of any empirical measure for the stability of a learning rule. Nevertheless, this also suggests two promising research directions; (*i*) a computationally efficient *mechanism* for choosing the value of $k$ to improve the reliability of the KFCV estimate; and (*ii*) a computationally efficient (meta)algorithm (hopefully with some guarantees) to estimate the stability of a learning rule.

On the theoretical side, our result, so far, is in terms of the stability of the learning rule $\mathtt{A}$, $k$, $n$, and $m$, but did not consider any particular learning rule in specific. As such, further insight and refined results can be obtained if our bound is applied to well known classes of algorithms such as potential function rules (which include the k-NN rule) [Devroye and Wagner, 1979], ridge regression, and binary classifiers for instance. Another interesting future work, which we hope could be built on our results, is to develop a posteriori generalization bounds for cross-validation estimation.

## Appendix

### Proof of Theorem 2

**Theorem 2.** *Let $Z$, $V_{DEL}$ be as in Corollary 1 and assume that $|Z - Z_{-i}| \leq 1$ almost surely for all $i$. For all $\theta > 0$, such that $\lambda \in (0, 1]$, $\theta\lambda < 1$, and $\mathbb{E}e^{\lambda V_{DEL}} < \infty$, the following holds:*

$$\log \mathbb{E}\left[\exp\left(\lambda(Z - \mathbb{E}Z)\right)\right] \leq \lambda\theta(1 - \lambda\theta)^{-1} \log \mathbb{E}\left[\exp\left(\lambda\theta^{-1}V_{DEL}\right)\right] \ . \tag{3}$$

*Proof.* The proof of this theorem relies on the result of Theorem 6.6 in [Boucheron et al., 2013] which we state here for convenience as a proposition without proof.

**Proposition 1.** *Let $\phi(u) = e^u - u - 1$. Then for all $\lambda \in \mathbb{R}$,*

$$\lambda\mathbb{E}\left[Z\exp(\lambda Z)\right] - \mathbb{E}\left[\exp(\lambda Z)\right]\log\mathbb{E}\left[\exp(\lambda Z)\right] \leq \sum_{i=1}^{n}\mathbb{E}\left[\exp(\lambda Z)\phi\left(-\lambda(Z - Z_{-i})\right)\right] \ . \tag{14}$$

To make use of inequality (14), we need to establish an appropriate upper bound for the RHS of (14). Note that for $u \leq 1$, $\phi(u) \leq u^2$. By assumption $|Z - Z_{-i}| \leq 1$ holds almost surely. Since $0 < \lambda \leq 1$, we get

$$\begin{aligned}\sum_{i=1}^{n}\mathbb{E}\left[\exp(\lambda Z)\phi\left(-\lambda(Z - Z_{-i})\right)\right] &\leq \lambda^2\sum_{i=1}^{n}\mathbb{E}\left[\exp(\lambda Z)\left(Z - Z_{-i}\right)^2\right] \\ &= \lambda^2\mathbb{E}\left[V_{\text{DEL}}\exp(\lambda Z)\right] \ .\end{aligned}$$

It follows that (14) can be written as

$$\lambda\mathbb{E}\left[Z\exp(\lambda Z)\right] - \mathbb{E}\left[\exp(\lambda Z)\right]\log\mathbb{E}\left[\exp(\lambda Z)\right] \leq \lambda^2\mathbb{E}\left[\exp(\lambda Z)V_{\text{DEL}}\right] \ . \tag{15}$$

The RHS of the previous inequality has two coupled random variables; $\exp(\lambda Z)$ and $V_{\text{DEL}}$. To make use of (14), we decouple the two random variables using the following useful tool from [Massart, 2000] which we state as a proposition without a proof.

**Proposition 2.** *For random variable $W$, and for any $\lambda \in \mathbb{R}$, if $\mathbb{E}\left[\exp(\lambda W)\right] < \infty$, then the following holds*

$$\frac{\mathbb{E}\lambda W\exp(\lambda Z)}{\mathbb{E}\exp(\lambda Z)} \leq \frac{\mathbb{E}\lambda Z\exp(\lambda Z)}{\mathbb{E}\exp(\lambda Z)} - \log\mathbb{E}\exp(\lambda Z) + \log\mathbb{E}\exp(\lambda W) \ . \tag{16}$$

Multiplying both sides of (16) by $\mathbb{E}\exp(\lambda Z)$ and replacing $W$ with $V_{\text{DEL}}/\theta$ we get that:

$$\mathbb{E}\exp(\lambda Z)V_{\text{DEL}} \leq \theta\left[\mathbb{E}Z\exp(\lambda Z) - \frac{1}{\lambda}\mathbb{E}\exp(\lambda Z)\log\mathbb{E}\exp(\lambda Z) + \frac{1}{\lambda}\mathbb{E}\exp(\lambda Z)\log\mathbb{E}\exp\left(\lambda\frac{V_{\text{DEL}}}{\theta}\right)\right] . \tag{17}$$

Introduce $F(\lambda) = \mathbb{E}\exp(\lambda Z)$, and $G(\lambda) = \log\mathbb{E}\exp(\lambda V_{\text{DEL}})$. Note that $F'(\lambda) = \mathbb{E}Z\exp(\lambda Z)$. Plugging (17) into (15) and using the compact notation $F(\lambda)$, $F'(\lambda)$, and $G(\lambda/\theta)$ we get that:

$$\lambda F'(\lambda) - F(\lambda)\log F(\lambda) \leq \lambda^2\theta\left(F'(\lambda) - \frac{1}{\lambda}F(\lambda)\log F(\lambda) + \frac{1}{\lambda}F(\lambda)G(\lambda/\theta)\right) \ . \tag{18}$$

Dividing both sides by $\lambda^2 F(\lambda)$ and rearranging the terms:

$$\frac{1}{\lambda}\frac{F'(\lambda)}{F(\lambda)} - \frac{1}{\lambda^2}\log F(\lambda) \leq \frac{\theta G(\lambda/\theta)}{\lambda(1 - \lambda\theta)} \ . \tag{19}$$

The rest of the proof continues exactly as the proof of Theorem 2 from [Boucheron et al., 2003]: As the left-hand side of the above display is just the derivative of $H(\lambda) = \frac{1}{\lambda}\log F(\lambda)$, (19) is equivalent to $H'(\lambda) \leq \frac{\theta G(\lambda/\theta)}{\lambda(1-\lambda\theta)}$. Recalling that $\lim_{\lambda\to 0+} H(\lambda) = \mathbb{E}[Z]$, the integration of the differential inequality gives $H(\lambda) \leq \mathbb{E}[Z] + \theta\int_0^\lambda \frac{G(s/\theta)}{s(1-s\theta)}\,ds$. Notice that $G$ is convex. This implies that the integrand is a nondecreasing function of $s$ and therefore $\log F(\lambda) \leq \lambda\mathbb{E}[Z] + \frac{\lambda\theta G(\lambda/\theta)}{1-\lambda\theta}$. □

**Proof of Lemma 1**

**Lemma 1.** *Let $Z$, $Z_{-i}$, $V_{DEL}$ be as in Theorem 2. If $V_{DEL} - \mathbb{E}V_{DEL}$ is a sub-gamma RV with variance parameter $v > 0$ and scale parameter $c \geq 0$, then for any $\delta \in (0,1)$, $a > 0$, with probability $1-\delta$,*

$$|Z - \mathbb{E}Z| \;\leq\; \tfrac{4}{3}(ac + 1/a)\log\left(\tfrac{2}{\delta}\right) + 2\sqrt{(\mathbb{E}V_{DEL} + a^2 v/2)\log\left(\tfrac{2}{\delta}\right)}\;.$$

*Proof.* Since $V_{\mathrm{DEL}} - \mathbb{E}V_{\mathrm{DEL}} \in \Gamma_+(v,c)$, for any $\lambda \in (0, 1/c)$ we have

$$\psi_{V_{\mathrm{DEL}} - \mathbb{E}V_{\mathrm{DEL}}}(\lambda) = \log \mathbb{E}\left[\exp(\lambda(V_{\mathrm{DEL}} - \mathbb{E}V_{\mathrm{DEL}}))\right] \leq \frac{\lambda^2 v}{2(1-c\lambda)}\;.$$

Rearranging the terms we get

$$\log \mathbb{E}\left[\exp(\lambda V_{\mathrm{DEL}})\right] \leq \lambda \mathbb{E}V_{\mathrm{DEL}} + \frac{\lambda^2(v/2)}{1-c\lambda}\;. \tag{20}$$

Combining this with the result of Theorem 2 where we choose $\theta = 1$, we get

$$\psi_{Z-\mathbb{E}Z}(\lambda) \leq \frac{\lambda}{1-\lambda}\left(\lambda\mathbb{E}V_{\mathrm{DEL}} + \frac{\lambda^2(v/2)}{1-c\lambda}\right)\;. \tag{21}$$

We upper bound the term on the right-hand side as follows:

$$\begin{aligned}
\frac{\lambda}{1-\lambda}\left(\lambda\mathbb{E}V_{\mathrm{DEL}} + \frac{\lambda^2(v/2)}{1-c\lambda}\right) &= \frac{\lambda}{1-\lambda}\left(\frac{\lambda\mathbb{E}V_{\mathrm{DEL}} - c\lambda^2\mathbb{E}V_{\mathrm{DEL}} + \lambda^2 v/2}{(1-c\lambda)}\right) \\
&\leq \frac{\lambda}{1-\lambda}\left(\frac{\lambda\mathbb{E}V_{\mathrm{DEL}} + \lambda^2(v/2)}{(1-c\lambda)}\right) \\
&= \frac{\lambda^2\mathbb{E}V_{\mathrm{DEL}} + \lambda^3(v/2)}{(1-\lambda)(1-c\lambda)} \\
&\leq \frac{\lambda^2\mathbb{E}V_{\mathrm{DEL}} + \lambda^2(v/2)}{(1-\lambda)(1-c\lambda)} \\
&= \frac{\lambda^2(\mathbb{E}V_{\mathrm{DEL}} + v/2)}{(1-\lambda)(1-c\lambda)} \\
&\leq \frac{\lambda^2(\mathbb{E}V_{\mathrm{DEL}} + v/2)}{(1-(c+1)\lambda)}\;,
\end{aligned}$$

where the last inequality holds provided that $0 < \lambda < 1/(c+1)$. Thus we finally get that

$$\psi_{Z-\mathbb{E}Z}(\lambda) \leq \frac{\lambda^2(\mathbb{E}V_{\mathrm{DEL}} + v/2)}{(1-(c+1)\lambda)}\;. \tag{22}$$

Recall that the Cramer-Chernoff method gives that for any $\lambda > 0$, $\mathbb{P}\left[Z > \mathbb{E}Z + t\right] \leq \exp(-(\lambda t - \psi_{Z-\mathbb{E}Z}(\lambda)))$. This combined with (22), we see that we need to lower bound $\lambda t - \psi_{Z-\mathbb{E}Z}(\lambda) \geq \lambda t - \frac{\lambda^2(\mathbb{E}V_{\mathrm{DEL}}+v/2)}{(1-(c+1)\lambda)}$, where $\lambda \in (0,1] \cap (0, 1/(c+1)) = (0, 1/(c+1))$ can be chosen so that the lower bound is the largest. From Lemma 11 in [Boucheron et al., 2003], we have that for any $p, q > 0$,

$$\sup_{\lambda\in[0,1/q)}\left(\lambda t - \frac{\lambda^2 p}{1-q\lambda}\right) \geq \frac{t^2}{4p + 2q(t/3)}\;,$$

and the supremum is attained at

$$\lambda = \frac{1}{q}\left(1 - \left(1 + \frac{qt}{p}\right)^{-1/2}\right)\;.$$

Setting $p = \mathbb{E}V_{\mathrm{DEL}} + v/2$, $q = c+1$, we see that the optimizing $\lambda$ belongs to $(0, 1/(c+1))$. Hence,

$$\mathbb{P}\left[Z > \mathbb{E}Z + t\right] \;\leq\; \exp\left(\frac{-t^2}{4(\mathbb{E}V_{\mathrm{DEL}} + v/2) + 2(c+1)t/3}\right)\;.$$

Letting the right hand side of the previous inequality to equal $\delta$ and solving for $t$ then after some further upper bounding to simplify the resulting expression (in particular, using $\sqrt{|a|+|b|} \le \sqrt{|a|} + \sqrt{|b|}$), we get

$$|Z - \mathbb{E}Z| \;\le\; \tfrac{4}{3}(c+1)\log\left(\tfrac{2}{\delta}\right) + 2\sqrt{(\mathbb{E}V_{\mathrm{DEL}} + v/2)\log\left(\tfrac{2}{\delta}\right)}\,. \tag{23}$$

The result now follows by applying (23) to $Z' = aZ$, $Z'_{-i} = aZ_{-i}$ and $V'_{\mathrm{DEL}} = \sum_i (Z' - Z'_{-i})^2$. Noting that $V'_{\mathrm{DEL}} = a^2 V_{\mathrm{DEL}} \in \Gamma(a^4 v, a^2 c)$, we get

$$a\,|Z - \mathbb{E}Z| \;\le\; \tfrac{4}{3}(a^2 c+1)\log\left(\tfrac{2}{\delta}\right) + 2\sqrt{(a^2\mathbb{E}V_{\mathrm{DEL}} + a^4 v/2)\log\left(\tfrac{2}{\delta}\right)}\,.$$

Dividing both sides by $a$ gives the desired inequality. □

**Proof of Lemma 2**

**Lemma 2.** *Let $Z$, $Z_{-i}$ be defined as in (8), and let $V_{DEL} = \sum_{i=1}^{k}(Z - Z_{-i})^2$. Then for any real $q \ge 1/2$, $k \ge 1$, and $n > m \ge 1$, the following holds:*

$$\|V_{DEL}\|_{2q} \;\le\; k\beta_{4q}^2(n-m,m)\,. \tag{9}$$

*Proof.* Let $q \ge 1$. Then,

$$\begin{aligned}
\|V_{\mathrm{DEL}}\|_q &= \left\|\sum_{i=1}^{k}(Z - Z_{-i})^2\right\|_q \\
&\le \sum_{i=1}^{k}\left\|(Z - Z_{-i})^2\right\|_q \qquad \text{(by triangle inequality)} \\
&= \sum_{i=1}^{k}\left\|\left(\frac{1}{k}\sum_{j=1}^{k}\widehat{R}\left(\mathtt{A}(\mathcal{S}_n^{-\mathcal{F}_j}), \mathcal{F}_j\right) - \frac{1}{k-1}\sum_{\substack{q=1\\ q\ne i}}^{k}\widehat{R}\left(\mathtt{A}(\mathcal{S}_n^{-\{\mathcal{F}_i,\mathcal{F}_q\}}), \mathcal{F}_q\right)\right)^2\right\|_q \\
&= \sum_{i=1}^{k}\left\|\left(\frac{1}{k(k-1)}\sum_{j=1}^{k}\sum_{\substack{q=1\\ q\ne i}}^{k}\left(\widehat{R}\left(\mathtt{A}(\mathcal{S}_n^{-\mathcal{F}_j}), \mathcal{F}_j\right) - \widehat{R}\left(\mathtt{A}(\mathcal{S}_n^{-\{\mathcal{F}_i,\mathcal{F}_q\}}), \mathcal{F}_q\right)\right)\right)^2\right\|_q \\
&\le \sum_{i=1}^{k}\left\|\frac{1}{k(k-1)}\sum_{j=1}^{k}\sum_{\substack{q=1\\ q\ne i}}^{k}\left(\widehat{R}\left(\mathtt{A}(\mathcal{S}_n^{-\mathcal{F}_j}), \mathcal{F}_j\right) - \widehat{R}\left(\mathtt{A}(\mathcal{S}_n^{-\{\mathcal{F}_i,\mathcal{F}_q\}}), \mathcal{F}_q\right)\right)^2\right\|_q \qquad \text{(by Jensen inequality)} \\
&\le \frac{1}{k(k-1)}\sum_{i=1}^{k}\sum_{j=1}^{k}\sum_{\substack{q=1\\ q\ne i}}^{k}\left\|\left(\widehat{R}\left(\mathtt{A}(\mathcal{S}_n^{-\mathcal{F}_j}), \mathcal{F}_j\right) - \widehat{R}\left(\mathtt{A}(\mathcal{S}_n^{-\{\mathcal{F}_i,\mathcal{F}_q\}}), \mathcal{F}_q\right)\right)^2\right\|_q \qquad \text{(by triangle inequality)} \\
&= \frac{1}{k(k-1)}\sum_{i=1}^{k}\sum_{j=1}^{k}\sum_{\substack{q=1\\ q\ne i}}^{k}\beta_{2q}^2(n-m,m) \\
&= k\beta_{2q}^2(n-m,m)\,.
\end{aligned} \tag{24}$$

Now observe that replacing $q$ with $2q$ yields that

$$\|V_{\mathrm{DEL}}\|_{2q} \le k\beta_{4q}^2(n-m,m)\,, \tag{25}$$

which completes the proof. □

**Proof of Lemma 4**

**Lemma 4.** *Let $Z$ and $Z_{-i}$ be defined as in (11) and let $V_{DEL} = \sum_{i=1}^{n}(Z - Z_{-i})^2$. Then for any real $q \geq 1/2$, and $n \geq 2$, the following holds:*

$$\|V_{DEL}\|_{2q} \;\leq\; n\beta_{4q}^2(n,1)\;. \tag{12}$$

*Proof.* Let $q \geq 1$. Then,

$$\begin{aligned}
\|V_{\text{DEL}}\|_q &= \left\|\sum_{i=1}^{n}(Z - Z_{-i})^2\right\|_q \\
&\leq \sum_{i=1}^{n}\left\|(Z - Z_{-i})^2\right\|_q \quad \text{(by triangle inequality)} \\
&= \sum_{i=1}^{n}\|(Z - Z_{-i})\|_{2q}^2 \quad \text{(since } \left\|X^2\right\|_q = \|X\|_{2q}^2\text{)} \\
&= \sum_{i=1}^{n}\left\|R(\mathtt{A}(\mathcal{S}_n),\mathscr{P}) - R(\mathtt{A}(\mathcal{S}_n^{-i}),\mathscr{P})\right\|_{2q}^2 \\
&= \sum_{i=1}^{n}\left\|\mathbb{E}\left[\ell\left(\mathtt{A}(\mathcal{S}_n),X\right) - \ell\left(\mathtt{A}(\mathcal{S}_n^{-i}),X\right)|\mathcal{S}_n\right]\right\|_{2q}^2 \\
&= \sum_{i=1}^{n}\left\|\mathbb{E}\left[\ell\left(\mathtt{A}(\mathcal{S}_n),X\right) - \ell\left(\mathtt{A}(\mathcal{S}_n^{-1}),X\right)|\mathcal{S}_n\right]\right\|_{2q}^2 \quad \text{(by } \textit{i.i.d} \text{ of the examples)} \\
&= n\beta_{2q}^2(n,1)\;.
\end{aligned} \tag{26}$$

Replacing $q$ with $2q$ yields that

$$\|V_{\text{DEL}}\|_{2q} \;\leq\; n\beta_{4q}^2(n,1)\;. \tag{27}$$

□

**Proof of Lemma 6**

**Lemma 6.** *Using the previous setup and definitions, let $\mathtt{A}$ be a learning rule with $L_2$ stability coefficient $\beta_2(n,m)$. Then, for $k \geq 1$, and $n > m \geq 1$, the following holds*

$$\left|\mathbb{E}R(\mathtt{A}(\mathcal{S}_n),\mathscr{P}) - \mathbb{E}\widehat{R}_{CV}(\mathtt{A},\mathcal{F}_{1,\ldots,k})\right| \;\leq\; \beta_2(n,m)\;.$$

*Proof.* To derive a bound on $|\mathbb{E}R(\mathtt{A}(\mathcal{S}_n),\mathscr{P}) - \mathbb{E}\widehat{R}_{\text{CV}}(\mathtt{A},\mathcal{F}_{1,\ldots,k})|$ in terms of $L_q$-stability, we proceed as follows. First, note that $\mathbb{E}R\left(\mathtt{A}(\mathcal{S}_n),\mathscr{P}\right) = \mathbb{E}\left[\ell\left(\mathtt{A}(\mathcal{S}_n),X\right)\right]$. Second, for $\mathbb{E}\widehat{R}_{\text{CV}}\left(\mathtt{A},\mathcal{F}_{1,\ldots,k}\right)$, we have

$$\begin{aligned}
\mathbb{E}\widehat{R}_{\text{CV}}\left(\mathtt{A},\mathcal{F}_{1,\ldots,k}\right) &= \mathbb{E}\left[\frac{1}{km}\sum_{j=1}^{k}\sum_{x_i\in\mathcal{F}_j}\ell\left(\mathtt{A}\left(\mathcal{S}_n^{-\mathcal{F}_j},x_i\right)\right)\right] \\
&= \frac{1}{km}\sum_{j=1}^{k}\sum_{x_i\in\mathcal{F}_j}\mathbb{E}\left[\ell\left(\mathtt{A}\left(\mathcal{S}_n^{-\mathcal{F}_j}\right),x_i\right)\right] \\
&= \frac{1}{km}\sum_{j=1}^{k}\sum_{i=1}^{m}\mathbb{E}\left[\ell\left(\mathtt{A}\left(\mathcal{S}_n^{-[m]}\right),x_i'\right)\right] \quad \text{(by } \textit{i.i.d} \text{ of the examples)} \\
&= \mathbb{E}\left[\ell\left(\mathtt{A}\left(\mathcal{S}_n^{-[m]}\right),X\right)\right]\;,
\end{aligned}$$

where $(X_1',\ldots,X_m')$ are *i.i.d* examples drawn from $\mathcal{X}$ according to $\mathscr{P}$.

It follows that

$$\begin{aligned}
\left|\mathbb{E}R\left(\mathtt{A}(\mathcal{S}_n), \mathscr{P}\right) - \mathbb{E}\widehat{R}_{\text{CV}}\left(\mathtt{A}, \mathcal{F}_{1,\ldots,k}\right)\right| &= \left|\mathbb{E}[\ell\left(\mathtt{A}(\mathcal{S}_n), X\right)] - \mathbb{E}[\ell(\mathtt{A}(\mathcal{S}_n^{-[m]}), X)]\right| \\
&= \left|\mathbb{E}\left[\ell\left(\mathtt{A}(\mathcal{S}_n), X\right) - \ell(\mathtt{A}(\mathcal{S}_n^{-[m]}), X)\right]\right| \\
&\leq \mathbb{E}\left[\left|\ell\left(\mathtt{A}(\mathcal{S}_n), X\right) - \ell(\mathtt{A}(\mathcal{S}_n^{-[m]}), X)\right|\right] \\
&\leq \sqrt{\mathbb{E}\left[\left(\ell\left(\mathtt{A}(\mathcal{S}_n), X\right) - \ell(\mathtt{A}(\mathcal{S}_n^{-[m]}), X)\right)^2\right]} \\
&= \beta_2(n, m). \quad \text{(by definition of } L_2 \text{ Stability)} \qquad (28)
\end{aligned}$$

□